\documentclass[journal]{IEEEtran}

\usepackage{amsmath,amssymb,amsfonts}
\usepackage{graphicx}
\usepackage{textcomp}
\usepackage{multicol}
\usepackage{subcaption}
\usepackage{graphicx}
\usepackage{multirow}
\usepackage{float}
\usepackage{cleveref}
\usepackage{algorithm2e}
\usepackage{url}

\usepackage[margin=0.4in]{geometry} 

\begin{document}

\title{\LARGE Building Trust in Conversational AI: A Comprehensive Review and Solution Architecture for Explainable, Privacy-Aware Systems using LLMs and Knowledge Graph}

\author{Ahtsham Zafar, Venkatesh Balavadhani Parthasarathy, Chan Le Van, Saad Shahid, Aafaq Iqbal khan, Arsalan Shahid}

\author{Ahtsham Zafar, Venkatesh Balavadhani Parthasarathy, Chan Le Van, Saad Shahid, Aafaq Iqbal khan, Arsalan Shahid

\thanks{All authors are with CeADAR - Ireland's Centre for Applied AI, University College Dublin, Dublin 4, Belfield, Ireland, Email: arsalan.shahid@ucd.ie
}}

\maketitle

\begin{abstract}

Conversational AI systems have emerged as key enablers of human-like interactions across diverse sectors. Nevertheless, the balance between linguistic nuance and factual accuracy has proven elusive. In this paper, we first introduce LLMXplorer, a comprehensive tool that provides an in-depth review of over 150 Large Language Models (LLMs), elucidating their myriad implications ranging from social and ethical to regulatory, as well as their applicability across industries. Building on this foundation, we propose a novel functional architecture that seamlessly integrates the structured dynamics of Knowledge Graphs with the linguistic capabilities of LLMs. Validated using real-world AI news data, our architecture adeptly blends linguistic sophistication with factual rigour, and further strengthens data security through Role-Based Access Control. This research provides insights into the evolving landscape of conversational AI, emphasizing the imperative for systems that are efficient, transparent, and trustworthy.

\end{abstract}


\IEEEpeerreviewmaketitle

\section{Introduction}
\label{sec:introduction}

The ability to comprehend and engage in natural language dialogue is an intrinsic aspect of human intelligence. In contrast, machines inherently lack this capability, necessitating methods to translate natural language into machine-readable formats and vice versa. Conversational AI, which has evolved since the mid-1960s, addresses this gap. Initial developments were based on pattern matching, while more recent approaches leverage Generative AI technologies for coherent, context-aware responses. This evolution has captured the interest of both researchers and the wider community, leading to applications across various domains such as healthcare \cite{healthcare1,healthcare2,healthcare3}, retail \cite{retail1,retail2}, journalism \cite{journalism1,journalism2}, finance \cite{finance1,finance2}, and education \cite{education1,education2}.

\subsection{History and Evolution of Conversational AI}

\begin{figure*}[hbt!]
    \centering
    \includegraphics[width=\textwidth]{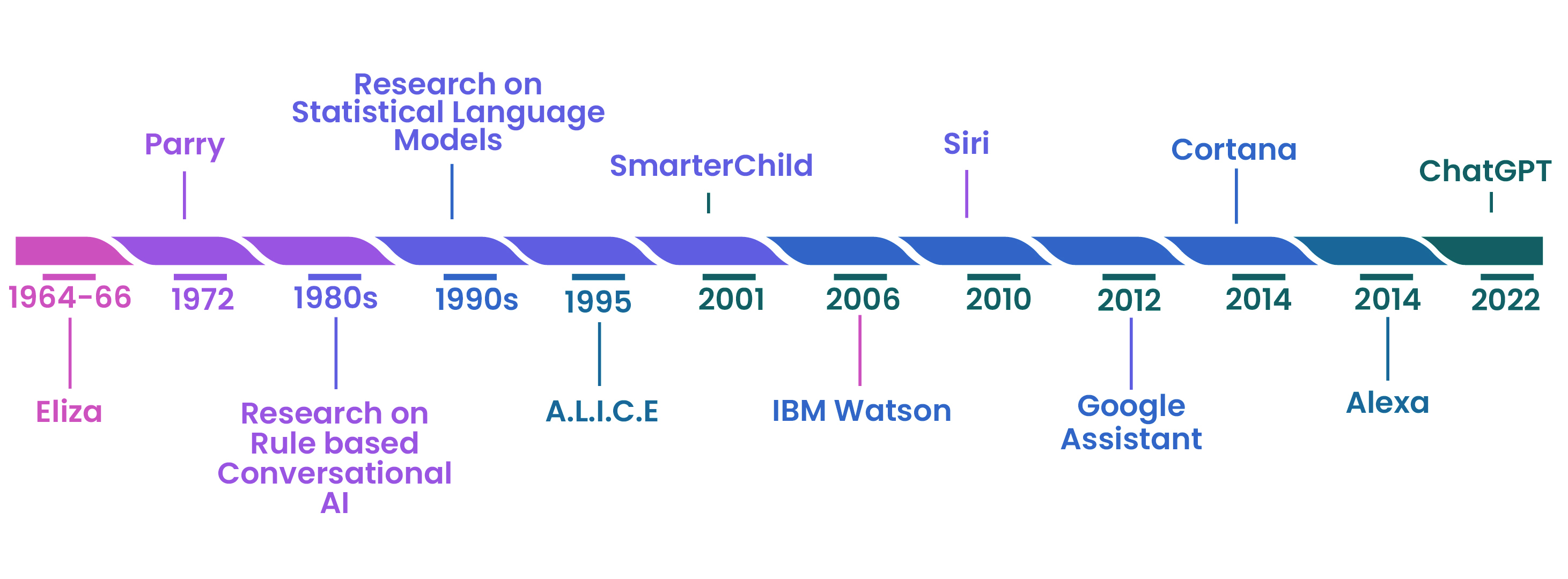}
    \caption{Timeline summarising the key product developments in conversational systems}
    \label{fig:ConversationalSystemsTL}
\end{figure*}

The journey of Conversational AI can be traced back to the mid-1960s, marked by seminal technologies like Eliza \cite{ELIZA}, PARRY \cite{PARRY}, and subsequent innovations such as Watson \cite{WATSON}, ALICE \cite{ALICE}, and GPT models \cite{GPT}. Eliza's keyword identification approach was succeeded by PARRY's conceptual modeling of mental disorders. A.L.I.C.E. in 1995 marked a significant shift by utilizing heuristic pattern matching with XML-based rules. The progression continued with SmarterChild's \cite{SmarterChild} natural language processing in 2001, IBM Watson's semantic analysis, and further personal assistants like Siri \cite{Siri}, Google Assistant \cite{GoogleAssistant}, Cortana \cite{Cortana}, and Alexa \cite{Alexa}. These developments culminated with OpenAI's ChatGPT \cite{ChatGPT}, grounded in the GPT-3.5 architecture, which harnessed vast datasets for more nuanced conversational capabilities. Figure \ref{fig:ConversationalSystemsTL} provides a timeline encapsulating these critical advancements.

\subsection{Development and Advancement of LLMs}

\begin{figure}[ht]
    \centering
    \includegraphics[width=9cm]{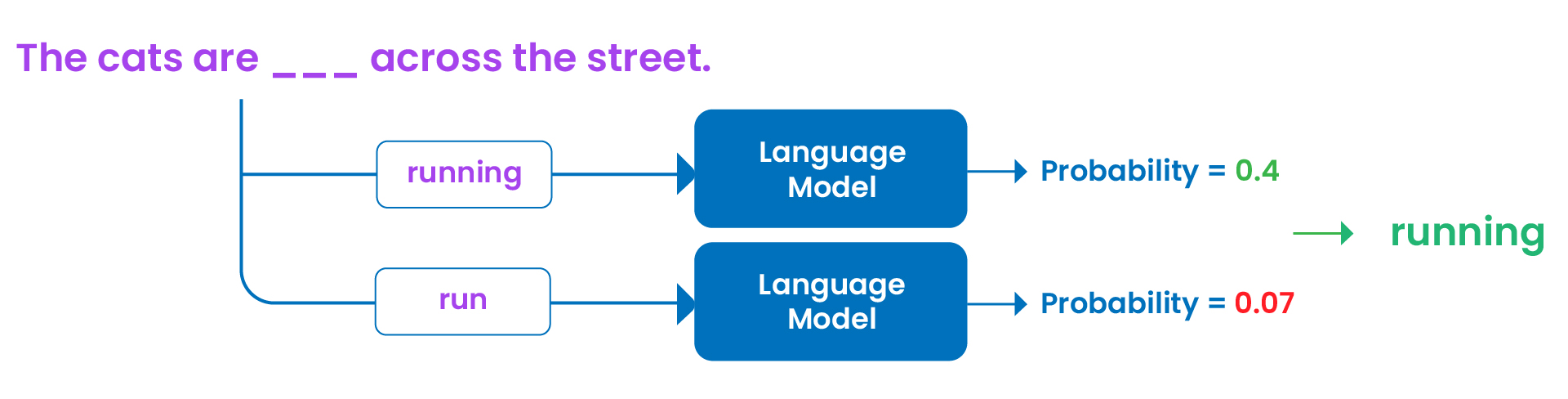}
    \caption{Illustration of LLM's prediction based on previous inputs}
    \label{fig:LLMPrediction}
\end{figure}

Large Language Models (LLMs) are critical components of machine learning, enabling the prediction of a word or character sequence within a given language. This can be formalised as predicting the \(N(i)\)-th word given the preceding sequence up to \(N(i-1)\) as illustrated in Figure \ref{fig:LLMPrediction}.

Historically, language models were constrained by limited training data, typically tailored for specific tasks like classification. The introduction of Generative Pre-trained Transformers (GPT) led to the evolution of Large Language Models (LLMs), expanding the modeling capabilities to encompass zero-shot \cite{zeroshot} and few-shot learning \cite{fewshot}. LLMs have proven to be versatile, with applications in various Natural Language Processing (NLP) tasks such as text summarization, machine translation, sentiment analysis, named entity recognition, etc.

The transformation began with BERT \cite{bert}, a pioneer in utilizing bidirectional encoding, simultaneously interpreting text from both directions. Following BERT, OpenAI's GPT-1 laid the groundwork for Generative AI but struggled with coherence. GPT-2 \cite{gpt2} marked significant advancements, introducing "Prompt Engineering" and achieving improvements in text generation. T5 \cite{t5} further expanded the field, leveraging an 11-billion-parameter model.

OpenAI's GPT-3 \cite{gpt3} became a landmark achievement, with 175 billion parameters and multilingual text generation. Google's PaLM \cite{palm}, featuring 540 billion parameters, and Meta's open-source LLaMA \cite{llama} continued this trend. OpenAI's ChatGPT, fine-tuned from GPT-3.5, specialized in dialogue generation through Reinforcement Learning with Human Feedback (RLHF).

Recent innovations include GPT-4 \cite{gpt4}, a multi-modal model capable of processing both text and images, although details remain scarce. The rapid advancements in LLMs have paved the way for more robust and versatile natural language understanding and generation, laying a solid foundation for the future of Conversational AI and related fields. \\

Despite the substantial advancements in LLMs and their growing prevalence in various applications, several inherent limitations and challenges remain. These include:

\begin{itemize}
    \item \textbf{Hallucination:} LLMs may generate information that is coherent but factually incorrect or misaligned with the underlying data.
    \item \textbf{Trustworthiness:} Ensuring the reliability and integrity of generated content poses a complex challenge.
    \item \textbf{Explainability:} The vast number of parameters and complex structures in LLMs often hinder clear understanding and interpretation of the models' decision-making processes.
    \item \textbf{Untraining Ability:} Modifying or retracting specific knowledge in an LLM without retraining the entire model remains an open challenge.
    \item \textbf{Privacy and Ethical Awareness:} Addressing concerns related to data privacy and ethical considerations in the application of LLMs requires careful scrutiny and novel approaches.
\end{itemize}

These challenges motivate the need for innovative and comprehensive solutions. This paper proposes a novel architecture that combines the strengths of Large Language Models (LLMs), Knowledge Graphs (KG), and Role-Based Access Control (RBAC). By integrating LLMs' natural language understanding capabilities with KG's structured representation of real-world knowledge and RBAC's access control and privacy mechanisms, the proposed architecture aims to address the aforementioned limitations. 

\subsection{Main Contributions and Paper Organisation}

The main contributions of this paper include:

\begin{itemize}
    \item \textbf{A Comprehensive Review of Large Language Models (LLMs):} We present a Large Language Model Explorer (LLMXplorer), an Excel-based tool that encapsulates over 150 key open and closed source LLMs. It provides a systematic overview of 16 key features, including parameters, training hardware, applications, industry relevance, country of origin, and more, offering insights into the state and evolution of the field.

    \item \textbf{Applied Analysis of LLMs in Various Industries:} An extensive investigation into the practical use cases, limitations, and challenges in technology adoption of LLMs across different sectors. This analysis contributes to a better understanding of the real-world applications and potentials for trustworthy conversational AI.

    \item \textbf{Functional Solution Architecture:} The development of a state-of-the-art, privacy-aware, explainable, and trustworthy conversational AI architecture. This design uniquely integrates LLMs, knowledge graph, and RBAC, and is empirically tested on the curated AI news dataset. 
\end{itemize}

Section~\ref{sec:llm-training} delves into the methodologies and training processes tailored for Large Language Models (LLMs), shedding light on key aspects such as architecture, training methodologies, and accessibility. In Section~\ref{sec:review_of_llm}, an exhaustive review of prominent LLMs is provided and LLMXplorer is introduced. Sections~\ref{sec:applied-implications-llms} and ~\ref{sec:market-analysis-llms} discuss the practical implications, technological impacts, market trends, and industry-specific applications of LLMs. Section~\ref{sec:sol-arch} introduces a functional solution architecture integrating LLMs with knowledge graphs, aiming for a trustworthy conversational AI system. Section~\ref{sec:discussion} presents the discussions and future implications. Finally, Section~\ref{sec:conclusion} concludes the paper.

\section{Methods and training process of LLMs} \label{sec:llm-training}

\begin{figure*}[hbt!]
\centering
\includegraphics[width=\textwidth]{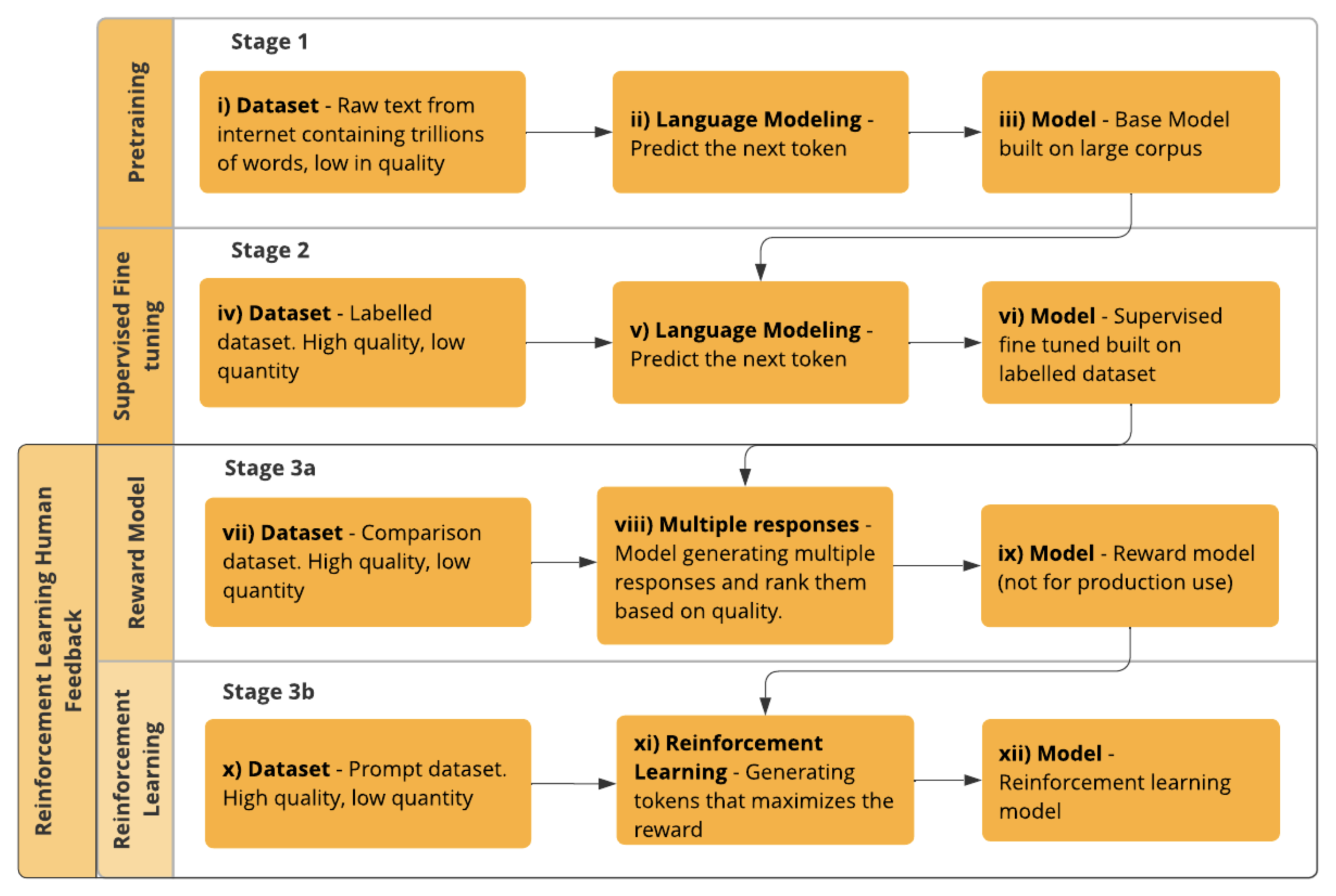}
\caption{Three-stage training process of Large Language Models (LLMs): Starting from an expansive pretraining on diverse data sources and utilizing the transformer architecture, transitioning into supervised fine-tuning with labeled datasets tailored for specific tasks, and culminating in dialogue optimization to refine AI-user interactions.}
\label{fig:LLMTraining}
\end{figure*}

Training a large language model is typically a multi-stage process. It consisted of two main stages: pretraining and supervised fine-tuning. Recent advancements have introduced a third stage, i.e., Reinforcement Learning from Human Feedback (RLHF). This stage has been observed to substantially enhance effectiveness for certain applications \cite{RLHF}. Figure \ref{fig:LLMTraining} depicts this tri-phase approach to LLM training.

\subsection{Pretraining}

During pretraining:

\begin{itemize}
    \item A vast amount of unlabeled data is gathered from diverse online sources, which might encompass trillions of tokens from articles, web pages, and books.
    \item Preprocessing procedures, such as tokenization, lowercasing, and special character removal, are applied. Tokenization might operate at the word or subword level.
    \item The popular choice for LLM training is the transformer architecture. This phase is computationally demanding, often requiring robust GPUs or TPUs and incurring significant time and cost.
\end{itemize}

\subsection{Supervised Fine-tuning}

Post pretraining, LLMs are fine-tuned using labeled datasets. They build upon the patterns discerned from pretraining, and refining for specific tasks. Notably, fine-tuning requires less data and computational power than building models anew. It has exhibited proficiency across a multitude of NLP tasks \cite{NLPtasks1, NLPtasks2, NLPtasks3}, owing to the marriage of broad linguistic knowledge from pretraining with task-focused adjustments.

\subsection{Dialogue Optimization}

The essence of dialogue optimization is refining the interaction between AI and users. The optimization has several facets:

\begin{itemize}
    \item \textbf{Language Understanding:} Amplifying the AI's prowess in grasping user inputs, encompassing intent detection and semantic interpretation.
    \item \textbf{Response Generation:} Enhancing the model's ability to craft context-aware, fluent replies, which might be diverse or user-tailored.
    \item \textbf{Context Management:} Upholding a consistent conversational context, encompassing dialogue history tracking and coherence maintenance.
    \item \textbf{Error Management:} Implementing mechanisms to detect and rectify mistakes or to address ambiguous inputs.
    \item \textbf{User Feedback:} Integrating user feedback loops to finetune and uplift the overall user experience.
\end{itemize}

\subsection{Reward Model}

Subsequent to supervised fine-tuning, the model produces an array of possible replies for a prompt. Human evaluators rank these based on quality. For further refinement, each response might be appended with a reward token. Although this process refines model outputs, it is labor-intensive and demands meticulous data curation.

\subsection{Reinforcement Learning from Human Feedback (RLHF)}

RLHF marries three steps: LLM pretraining, human preference-based reward model training, and reinforcement learning-guided LLM fine-tuning. Despite its efficacy, RLHF has its challenges, including the cost of obtaining human feedback and potential risks associated with producing detrimental or false content.

\section{Comprehensive review of state-of-the-art LLMs} \label{sec:review_of_llm}

The October 2022 release of OpenAI's GPT3.5~\cite{GPT3.5} marked a significant milestone in the evolution of Large Language Models (LLMs). Subsequently, there was a noticeable uptick in the development of both open and closed-source models. Notably, a slew of open-source initiatives sought to replicate or surpass GPT3.5's capabilities. Renowned entities such as Google, Meta, Nvidia, Huawei, and Microsoft have unveiled LLM-based services, harnessing generative AI to augment their portfolios. The primary goal of these endeavours is the broadened accessibility of LLMs. This section gleans insights from an exhaustive review of recent literature on LLMs.

\subsection{LLMXlorer: Large Language Model Explorer}

With the proliferation of open-source LLMs, there are now over 16,000 distinct variations~\cite{llms-scrapped}. The Large Language Model Explorer (LLMXplorer)\footnote{LLMXplorer is freely available. Access requests can be sent via: \url{https://forms.gle/TNUbqHiCBsinD4Bu8}} ~\cite{zafar2023llmxplorer} has been introduced as an Excel-based tool, providing insights on a concise catalogue of over 150 LLMs based on strict inclusion criteria such as public accessibility, detailed documentation, practical applications, recognizable releasing entities, and comprehensive training details. It classifies them based on capabilities, applications, industry sectors, and more. The LLMXplorer captures 17 key attributes for both open-source and proprietary models including:

\begin{itemize}
\item \textit{Release Date:} Official launch date of the LLM.
\item \textit{Number of Parameters:} Indication of model complexity.
\item \textit{Base Model:} Foundational pre-trained model for the LLM.
\item \textit{Training Data Size:} Dataset volume used in training.
\item \textit{Training Hardware:} Infrastructure used in model training.
\item \textit{Training Time:} Duration of model training.
\item \textit{Context length:} Length of input text the model can process in one go.
\item \textit{Target Application:} Primary use-case of the LLM.
\item \textit{Model License Type:} Governing license of the LLM.
\item \textit{Modalities:} Supported data types (text, image, video).
\item \textit{Releasing Company:} Entity releasing the LLM.
\item \textit{Industry:} Targeted domain or sector of the LLM.
\item \textit{Origin Country:} Base country of the releasing entity.
\item \textit{Estimated User Base:} Approximate adoption rate.
\item \textit{Privacy Awareness:} Model's emphasis on data protection.
\item \textit{Ethical Awareness:} Model's ethical design and use considerations.
\item \textit{Reference Link:} Source link for more LLM details.
\end{itemize}

\subsection{Insights from LLMXplorer} 

\begin{figure}[hbt!] 
    \centering
    \includegraphics[width=9cm]{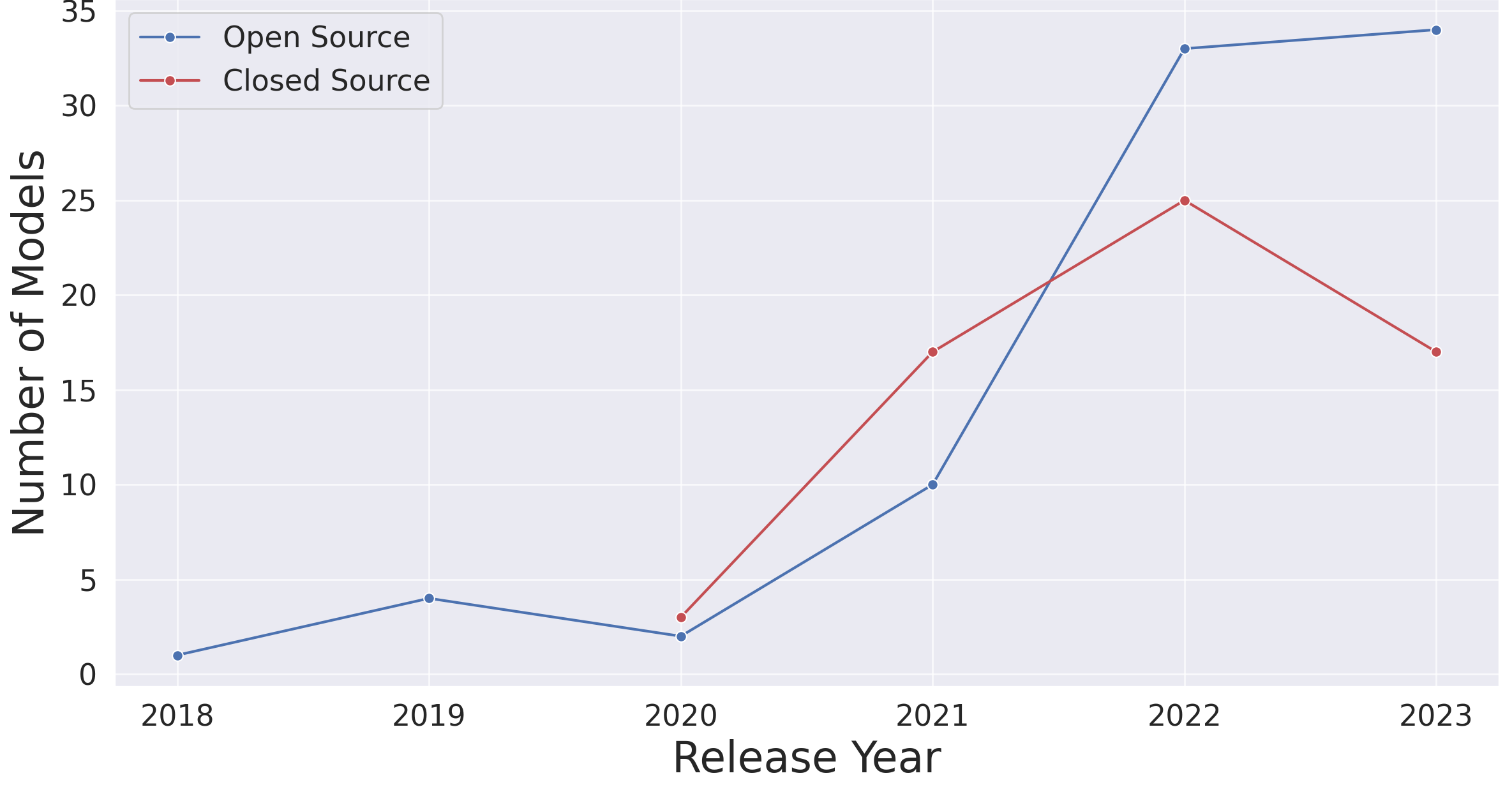}
    \caption{Comparative analysis of annual releases: Open-source (n=89) vs. Closed-source (n=64) LLMs.}
    \label{fig:number-of-LLMs}
\end{figure}

\begin{figure*}[hbt!] 
    \centering
    \includegraphics[width=15cm]{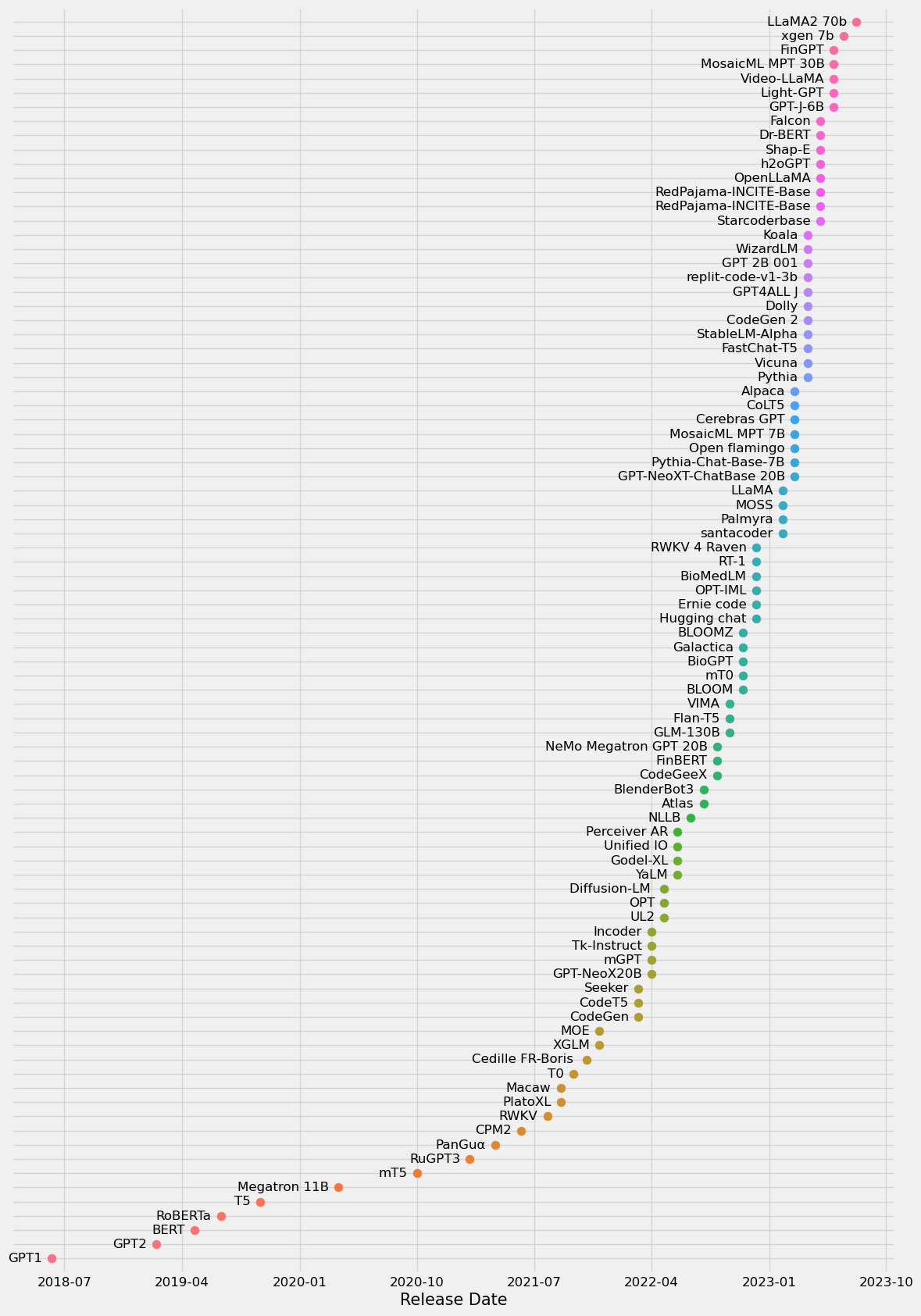}
    \caption{Timeline of Open Source LLMs (n=89). The x-axis displays year-month, while the y-axis shows vertical stacks depicting the total number of models (in each stack) released within one month.}
    
    \label{fig:Open Source LLMs}
\end{figure*}

\begin{figure*}[hbt!] 
    \centering
    \includegraphics[width=13cm]{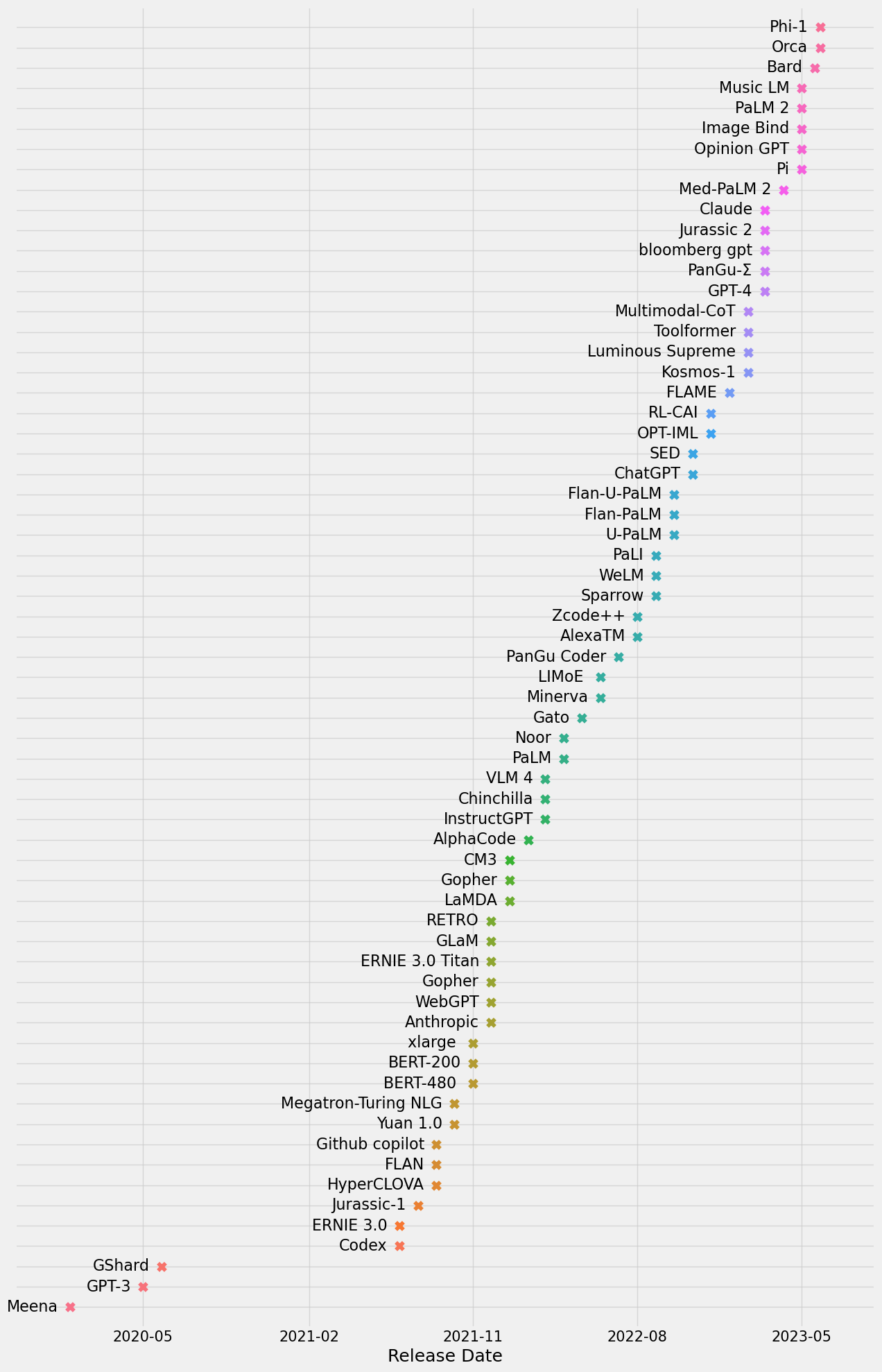}
    \caption{Timeline of Closed Source LLMs (n=64). The x-axis displays year-month, while the y-axis shows vertical stacks depicting the total number of models (in each stack) released within one month.}
    \label{fig:Closed Source LLMs}
\end{figure*}

\begin{figure}[hbt!] 
    \centering
    \includegraphics[width=9cm]{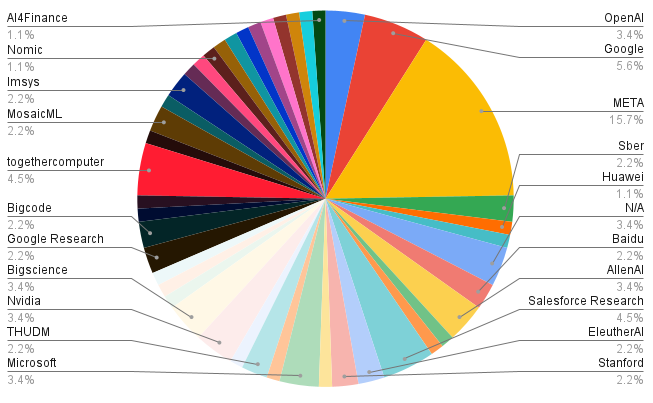}
    \caption{Open Source LLMs developed by companies (n=89). The chart illustrates the percentage distribution of models developed by various companies.}
    
    \label{fig:Open Source by company}
\end{figure}

\begin{figure}[hbt!] 
    \centering
    \includegraphics[width=9cm]{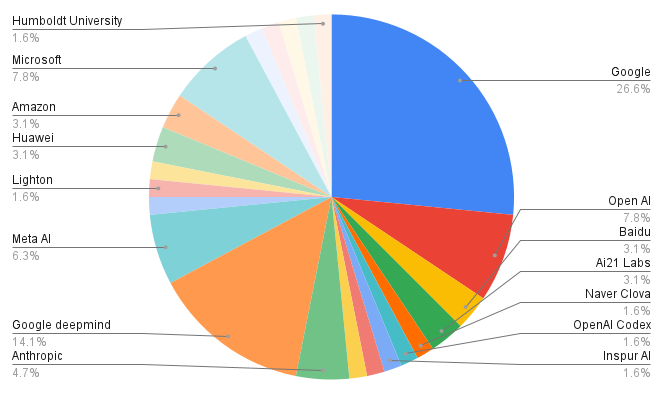}
    \caption{Closed Source LLMs developed by companies (n=64). The chart illustrates the percentage distribution of models developed by various companies.}
    \label{fig:Closed source by company}
\end{figure}

\begin{figure}[hbt!] 
    \centering
    \includegraphics[width=9cm]{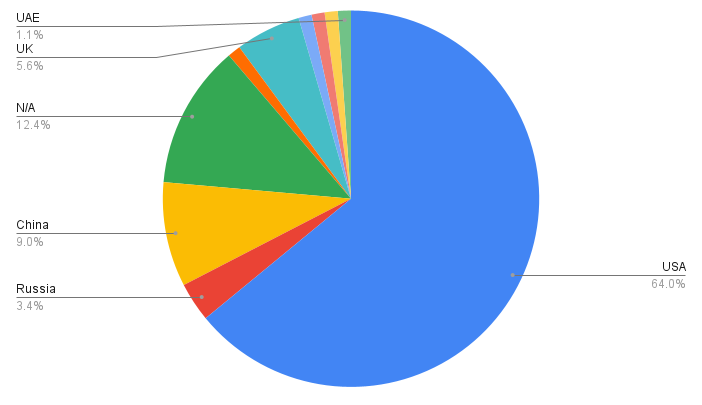}
    \caption{Open Source LLMs categorized by the originating country of the developer company (n=89). The pie chart segments are labeled with the respective country names and their corresponding percentage contributions. }
    \label{fig:Open source by country}
\end{figure}

\begin{figure}[hbt!] 
    \centering
    \includegraphics[width=9cm]{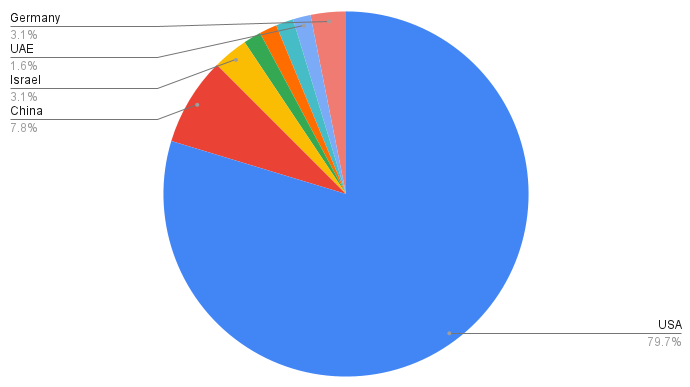}
    \caption{Closed Source LLMs categorized by the originating country of the developer company (n=64). The pie chart segments are labeled with the respective country names and their corresponding percentage contributions.}
    \label{fig:Closed source by country}
\end{figure}

\begin{figure}[hbt!] 
    \centering
    \includegraphics[width=9cm]{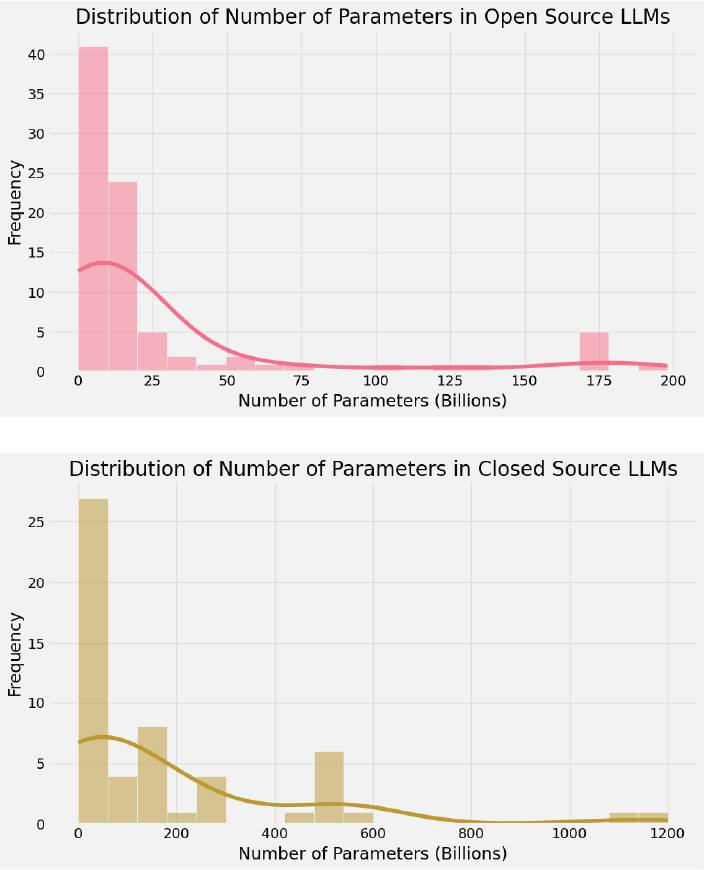}
    \caption{Distribution of Number of Parameters (in billions) in Open and Closed Source LLMs (n=89 for Open Source, n=64 for Closed Source). The x-axis depicts the parameter count in billions (B), while the y-axis represents the frequency (i.e., the number of models within each histogram bin). The solid line (KDE) illustrates the kernel density estimate, offering a smoothed distribution shape insight.}
    \label{fig:open-closed-parameters}
\end{figure}

\begin{itemize}
    \item Figures~\ref{fig:Open Source LLMs} and \ref{fig:Closed Source LLMs} detail LLM development timelines for open and closed-source categories, respectively. From Figure~\ref{fig:number-of-LLMs}, in the past six years, open-source LLMs averaged 14 annual releases with a standard deviation of 15.4. The annual maximum was 34 models. For closed-source LLMs, over four years, the average is 15.5 releases per year with a standard deviation of 9.15. The range spans from 3 to 25 models annually. Both categories exhibit a rising trend in annual releases, signaling increased LLM development interests. Notably, closed-source models show a slower growth rate, possibly indicating varied community involvement or resource allocation.
    \item Figures~\ref{fig:open-closed-parameters} depict parameter distributions for open and closed-source models. Open-source LLMs range from 0.035 billion to 198 billion parameters, averaging 27.77 billion. The models CPM2, BLOOM, BLOOMZ, OPT, and BlenderBot3 exceed 175 billion parameters. Closed-source LLMs range from 0.06 billion to 1200 billion, averaging 182.94 billion. Notably, GLaM and PanGu-E surpass 1 trillion parameters.
    \item Figures \ref{fig:Open source by country} and \ref{fig:Closed source by country} show LLM development distribution by country. For open-source models, the USA dominates, followed by the UK, Canada, and China. A similar distribution is observed for closed-source models.
    \item Figures \ref{fig:Open Source by company} and \ref{fig:Closed source by company} represent LLM development frequencies by leading companies. META dominates open-source development, whereas Google Research leads in closed-source LLMs.
\end{itemize}

\section{Applied and Technology Implications for LLMs} \label{sec:applied-implications-llms}

This section highlights the social, ethical, legal, privacy, regulatory, and physiological implications of LLMs.

\subsection{Social and Ethical Implications}

\textit{Democratization of Information:} LLMs' open-source availability has broadened AI access, spurring innovation among diverse users \cite{democratize}. Despite the slight quality lead of tech giants, open-source models present compelling alternatives, shifting the industry's landscape. Notably, personal LLMs, like LoRA \cite{lora}, allow efficient internal data handling and quick model fine-tuning.

\textit{Bias and Fairness:} Complexity in LLMs introduces misuse and bias risks. Addressing these necessitates diverse development teams and thorough bias audits to ensure fairness.

\textit{Digital Divide:} Disparities in LLM access can widen societal digital divides. Addressing this requires promoting technology access and LLM education \cite{digitaldivide}.

\textit{Responsible AI Development:} Balancing LLM benefits against risks demands collaboration among varied stakeholders. Transparency, robust data governance, and educational initiatives can shape responsible LLM use.

\textit{Algorithmic Transparency:} The "black box" nature of LLMs challenges trust and understanding. Research into LLM explainability and user-friendly output interpretations is essential.

\textit{Mitigating Harmful Content:} Ensuring diverse training data and implementing watermarking can curb harmful LLM outputs. Oversight mechanisms, both algorithmic and human, can bolster responsible LLM deployment \cite{mitigateharmful}.

\subsection{Legal, Privacy, and Regulatory Perspective}

\textit{Data Protection and Privacy Laws:} LLMs must adhere to regulations like GDPR while preventing unintentional data disclosure. Recent regulatory shifts, like the "EU AI ACT" \cite{EUact} and proposed Chinese regulations, emphasize responsible generative AI use.

\textit{Intellectual Property Considerations:} LLMs' capacity to generate human-like text challenges copyright norms. Addressing this requires clear intellectual property guidelines and plagiarism detection tools \cite{IP}.

\subsection{Physiological Perspectives}

\textit{Human-Computer Interaction:} LLM-based Conversational Agents carry potential risks, including undue trust, psychological vulnerabilities, and design biases. Promoting responsible design and prioritizing user privacy is paramount.

\begin{itemize}
\item Anthropomorphized systems risk user overreliance, potentially yielding unsafe outcomes or psychological harm. Emphasizing oversight and responsible use can counteract these threats.
\item Users might inadvertently trust conversational agents with private information. Protecting against misuse demands robust safeguards and user awareness campaigns.
\item Design choices in conversational agents can inadvertently perpetuate stereotypes. Addressing these requires design sensitivity and unbiased principles \cite{HCI}.
\end{itemize}

\section{Market Analysis of LLMs and Cross-Industry Use cases} \label{sec:market-analysis-llms}

\subsection{Market Size of LLMs and Driving Factors}

The global Natural Language Processing (NLP) market is projected to surpass 91 billion USD by 2030, expanding at a CAGR of 27\% \cite{straightsresearch}. A major driving force behind this growth is the influence of Large Language Models (LLMs). While exact market figures for LLMs remain elusive, certain factors listed in Table~\ref{tab:market-size-factors} hint at their increasing dominance. Despite the absence of precise market data, it is evident that the sector demonstrates substantial potential and is set to continue expanding across various industries.

\begin{table}[hbt!]
\footnotesize
\renewcommand{\arraystretch}{1.2}
\caption{Factors driving the LLM market growth}
\label{tab:market-size-factors}
\begin{tabular}{p{2.5cm} p{5.1cm}}
\hline
\textbf{Factors} & \textbf{Description} \\
\hline
Enterprise Adoption & Organizations utilize LLMs to optimize customer service, automate operations, and enhance efficiency. Sectors such as e-commerce, healthcare, and finance are primary beneficiaries. \\
\hline
AI Industry Growth & LLMs, central to AI initiatives, benefit from the holistic expansion of the artificial intelligence sector, accentuated by deep learning advancements. \\
\hline
Language-as-a-Service (LaaS) Emergence & LaaS platforms provide ready-to-use language models and APIs, enabling enterprises to harness language capabilities without intensive model training. \\
\hline
R\&D Momentum & Continuous advancements in NLP and machine learning propel the market. Innovations in model design, training techniques, and performance optimization catalyze industrial interest. \\
\hline
\end{tabular}
\end{table}

\subsection{LLM Development Opportunities}

Key applied challenges in the development of LLMs include:

\begin{itemize}
    \item \textit{Disinformation Generation}: LLMs can generate convincing misinformation, undermining information credibility and leading to potential harm. 
    \item \textit{Deepfakes Creation}: LLMs can produce sophisticated manipulated media, posing threats of deception and public manipulation. 
    \item \textit{Privacy and Data Concerns}: Handling vast training data introduces privacy vulnerabilities, potentially culminating in breaches and unintentional sensitive data exposure. 
    \item \textit{Bias Amplification}: LLMs might reflect and amplify biases present in training datasets, potentially reinforcing societal prejudices. 
    \item \textit{Malicious Usage}: LLMs can be tailored to produce harmful, offensive, or abusive content, potentially instigating hate or targeted assaults. 
    \item \textit{Unintended Outcomes}: Outputs from LLMs can occasionally have unexpected ramifications or ethical concerns, underscoring the need for careful oversight and guidelines. 
\end{itemize}

Recognizing the challenges associated with LLMs, there exists a spectrum of opportunities for refining and evolving these models (Table~\ref{tab:LLM-Opportunities}).

\begin{table}[hbt!]
\footnotesize
\renewcommand{\arraystretch}{1.2}
\caption{Opportunities in LLM Evolution}
\label{tab:LLM-Opportunities}
\begin{tabular}{p{2.5cm} p{5.1cm}}
\hline
\textbf{Opportunity} & \textbf{Description} \\
\hline
Bias and Fairness & Develop techniques to detect and neutralize biases in LLMs, ensuring outputs that respect diverse perspectives. \\
\hline
Ethical Guidelines & Forge ethical standards and frameworks for the responsible creation and deployment of LLMs. \\
\hline
Explainability & Innovate methods to augment LLM transparency and interpretability, clarifying decision-making processes for users. \\
\hline
Contextual Accuracy & Enhance LLMs' contextual comprehension to produce more relevant and precise responses. \\
\hline
Collaborative Efforts & Foster collaborations amongst stakeholders, like researchers and policymakers, to address LLM-related societal and ethical implications. \\
\hline
\end{tabular}
\end{table}

\subsection{Technology Applications and Use Cases in Diverse Industrial Sectors}

At present, the broad application potential of LLMs remains partially untapped due to concerns regarding data sensitivity and model ethics. However, the evolution of privacy-aware LLM models could redefine numerous industrial landscapes. A selection of potential applications across various sectors is elucidated in Appendix~\ref{app:usecases}. These include finance and risk management (Table~\ref{tab:llm-applications-finance}), healthcare and diagnostics (Table~\ref{tab:llm-applications-healthcare}), education and e-learning (Table~\ref{tab:llm-applications-education}), customer service and support (Table~\ref{tab:llm-applications-customer-service}), marketing and sales (Table~\ref{tab:llm-applications-marketing-sales}), human resources and recruitment (Table~\ref{tab:llm-applications-hr}), legal and compliance (Table~\ref{tab:llm-applications-legal}) and supply chain management (Table~\ref{tab:llm-applications-supply}). Notably, these listed applications represent a mere fraction of the extensive possibilities domain-specific LLMs can offer.

\section{Solution Architecture for Privacy-aware and Trustworthy Conversational AI} \label{sec:sol-arch}

With the advancement of conversational AI platforms powered by LLMs, there arises a need for addressing issues of explainability, data privacy, and ethical use of data to ensure their safe usage in industry and practical applications. LLMs, despite their significant natural language understanding capabilities, often operate as black-box models, making their decisions difficult to interpret. Integrating Knowledge Graphs (KGs)~\cite{Fensel2020} with LLMs offers a solution to these challenges by coupling the structured knowledge representation of KGs with the linguistic proficiency of LLMs. Furthermore, the incorporation of Role-Based Access Control (RBAC) into the architecture ensures that access to the AI system aligns with organizational policies and is granted only to authorized roles.

\textbf{Advantages of Key Components:}

\begin{itemize}
    \item \textbf{LLMs:} 
    \begin{itemize}
        \item Offer deep linguistic understanding, facilitating nuanced interactions.
        \item Capable of continuous adaptation based on user interactions.
        \item Provide a foundational layer for conversational platforms.
    \end{itemize}
    
    \item \textbf{KGs:} 
    \begin{itemize}
        \item Deliver structured and validated domain-specific knowledge.
        \item Enhance system explainability by tracing the origin of information.
        \item Complement LLMs by addressing gaps in domain-specific expertise.
    \end{itemize}
    
    \item \textbf{RBAC:} 
    \begin{itemize}
        \item Ensure data privacy through controlled access mechanisms.
        \item Align AI system usage with organizational policies and compliance mandates.
        \item Offer a methodical approach to managing data access, advancing security.
    \end{itemize}
\end{itemize}

Several architectural paradigms can merge KG and LLM. As detailed by Shirui Pan et al. in \cite{kg-LLM-paper}, the primary techniques include KG-enhanced LLMs, LLM-augmented KGs, and a synergistic LLMs+KGs approach. Our proposition of a closed-loop architecture is in line with this synergistic design, underscoring the advantages of combining LLMs and KGs. 

\subsection{System Components and Design Choices}

To elucidate the practical implementation of our architectural design, we anchor our exposition using media and journalism, specifically leveraging news data aggregated from the AI NewsHub platform~(\url{https://ainewshub.ie}). 

\subsubsection{AI NewsHub Dataset Description}

The AI NewsHub dataset is a systematically aggregated collection of news articles, procured daily from web crawling and search engines, and stored within a relational database system. Each record within this dataset represents an article and encapsulates the following attributes: identifier (ID), title, article content, published date, publisher name, and associated country of origin. Upon acquisition, articles undergo an analytical phase wherein they are classified based on their relevant topics, affiliated industry sectors, and originating publishers' categories.

\subsubsection{Construction of the AI NewsHub-based KG:}

The structural organization of KG facilitates the extraction of nuanced insights, supports decision-making processes, and aids in the identification of concealed patterns within the data.

The KG derived from the AI NewsHub dataset comprises two primary node classes: articles and topics. The 'article' node encompasses attributes such as ID, content, sentiment analysis results, and a multi-dimensional matrix or vector array. 

The former two attributes are directly extracted from the article dataset, while the sentiment is derived via the Vader sentiment analysis tool\cite{vader}, post preprocessing of the article content. The vector array is algorithmically generated leveraging the FastText library\cite{fasttext}, contingent upon the content of the article. 

Simultaneously, an 'article' node establishes a directed relationship with the 'topic' node when it is categorized under a specific topic. This 'topic' node bears the topic ID and the topic name, extracted from the designated topic table in the dataset. Storage and management of this KG are facilitated through Neo4j, with queries being structured using the Cypher language.

\textbf{Neo4j and Cypher:}

Neo4j~\cite{neo4j}, a premier graph database management platform, is adept at administering and querying intricate KGs. Given our emphasis on discerning the intricate interconnectedness of data, Neo4j is an optimal and open-source choice. In tandem with Cypher~\cite{cypher} — its dedicated query language — Neo4j empowers us to model multifaceted relationships, navigate patterns, and distil valuable insights from the KG. Cypher's design emphasizes human-readable syntax, ensuring that users can structure and interpret complex queries with relative ease.

\subsubsection{Llama-2 LLM}

In July 2023, Meta introduced the Llama-2 series~\cite{llama2}, an advancement in its Large Language Model lineup. The models in this series are characterized by parameters ranging from 7 billion to 70 billion. When compared to the preceding Llama-1 models, Llama-2 variants have been trained on an expanded dataset with an increase in token count by 40\% and an extended context length of 4,000 tokens. The 70B model incorporates the grouped-query attention mechanism~\cite{GQA}, which is designed for efficient inference.

A specialized variant within the Llama-2 series is the Llama-2-Chat, fine-tuned for dialogue applications using Reinforcement Learning from Human Feedback (RLHF). Benchmark evaluations indicate that the Llama-2-Chat model offers improvements in areas such as user assistance and system security. Performance metrics suggest that this model's capabilities align closely with those of ChatGPT, as per human evaluators. Additionally, the 70B version of Llama-2 has been recognized on HuggingFace’s OpenLLM leaderboard, exhibiting strong performance on several benchmarks, including but not limited to ARC~\cite{arc}, HellaSwag~\cite{hellaswag}, MMLU~\cite{MMLU}, and TruthfulQA (MC)~\cite{truthfulqa}.

With the details of the three main components provided, we will now describe the proposed system architecture in the next section.

\subsection{Architecture workflow}

\begin{figure*}
    \centering
    \includegraphics[width=\textwidth]{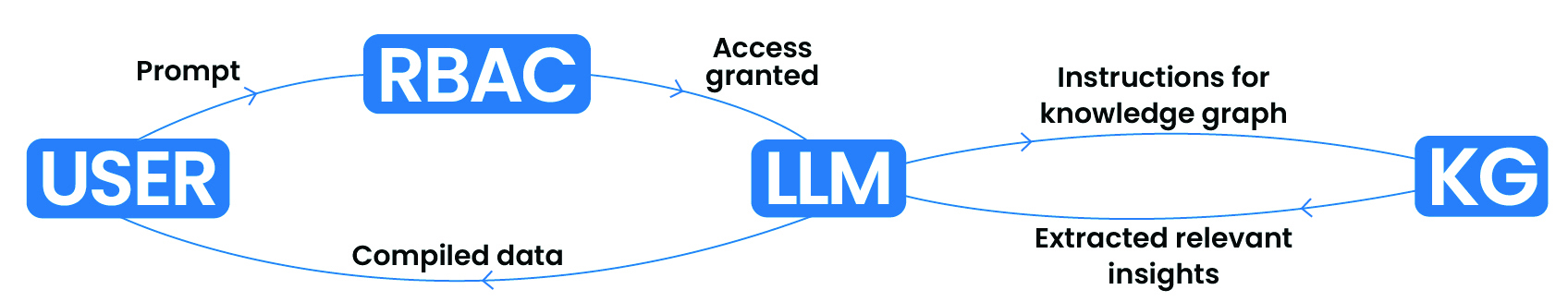}
    \caption{Simplified functional workflow for combined Knowledge Graph and LLM}
    \label{fig:simplified workflow}
\end{figure*}

\begin{figure*}
    \centering
    \includegraphics[width=0.95\textwidth]{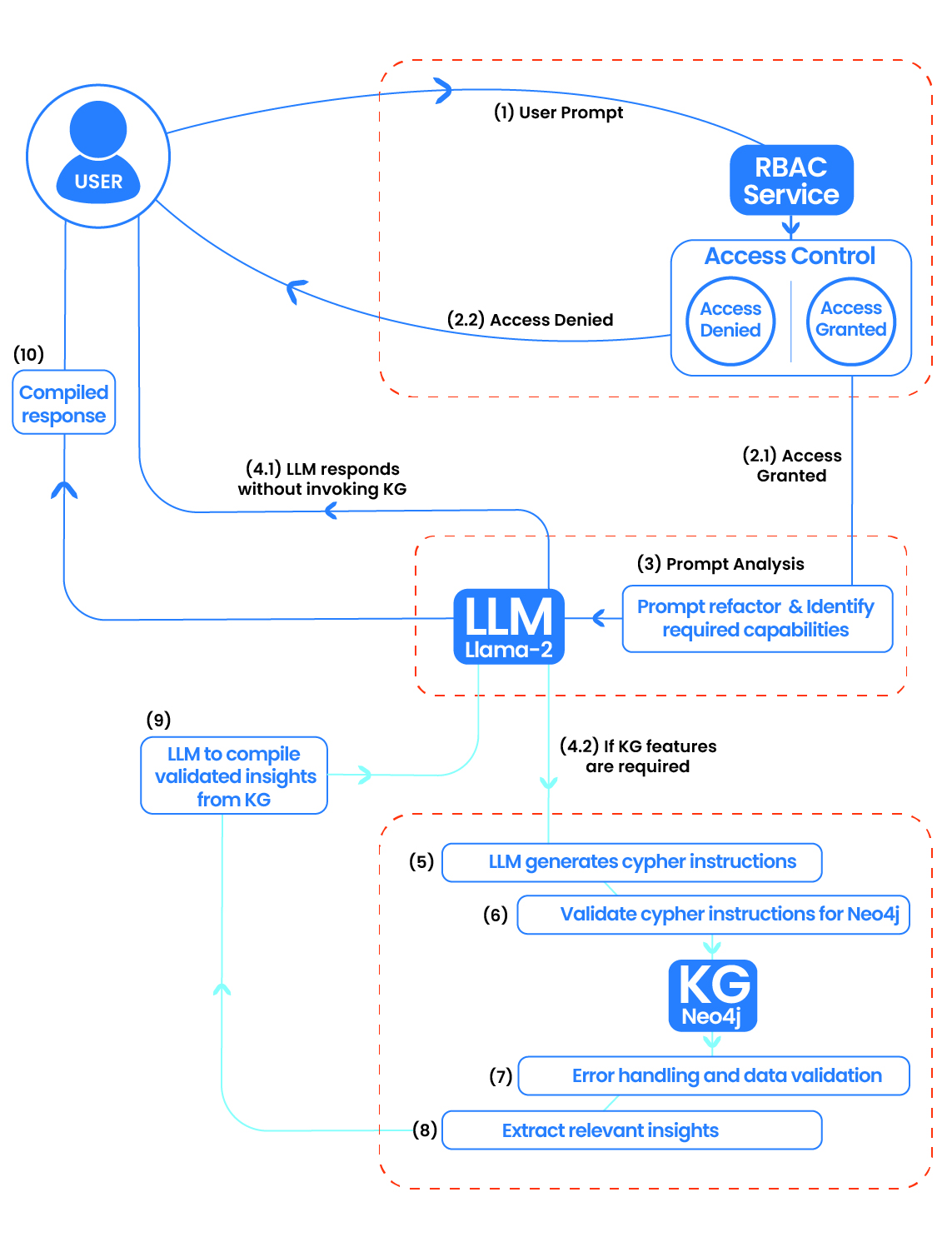}
    \caption{Functional architecture illustrating the integration of Large Language Models (LLMs), Knowledge Graphs (KG), and Role-Based Access Control (RBAC).}
    \label{fig:KG-LLM-detailed-workflow}
\end{figure*}

Figures~\ref{fig:simplified workflow} and~\ref{fig:KG-LLM-detailed-workflow} present the architectural workflow of our proposed system. The former offers a more general overview, elucidating the synergized connection between these three components. The latter is a detailed view of the intricate interactions between Knowledge Graphs (KG), Large Language Models (LLM), and Role-Based Access Control (RBAC).

The detailed sequence of operations in the system is described as follows:

\begin{itemize}
    \item \textbf{Step 1}: The User (U) communicates a specific request to the RBAC Service (S).

    \item \textbf{Step 2}: The RBAC Service (S) evaluates the permissions associated with the user, forwarding the request to the Access Control (AC). AC determines the user's data access rights. Upon granting access, the process proceeds to \textit{Step 3}. If denied, it transitions to \textit{Step 2.2}.

    \item \textbf{Step 3}: The Prompt Analysis module refines and refactors the user's prompt (if necessary) and identifies the key capabilities required to put together an appropriate response. In our case of journalism, the key capabilities include natural language understanding and generic output response or specialised capabilities such as similar article finder, sentiment analysis, fact-checking, and prediction for article topics and relevant industry sectors.

    \item \textbf{Step 4}: The Llama-2 LLM processes the user request based on the identified capabilities. If a generic response is required, the LLM responds to the user directly (Step 4.1). If specialised features from KG are required, the process moves to Step 4.2. 

    \item \textbf{Step 5}: Llama-2 generates or invokes relevant Cypher instructions for Neo4j based on required capabilities. 
    
    \item \textbf{Step 6}: A Cypher Validation Layer (CVL) ensures the integrity and safety of these instructions. Validated queries are executed on the Neo4j Knowledge Graph.

    \item \textbf{Step 7}: KG processes the queries, extracting the pertinent data. An Error Handling (EH) mechanism inspects the extracted insights for anomalies.

    \item \textbf{Step 8}: Error-free insights are compiled to be returned back to the LLM.
    
    \item \textbf{Step 9 and 10}: LLM formats the insights for user-friendly presentation and provides a response to the user. The User (U) receives the curated data, ensuring only permitted information is accessed. Users have the option to offer feedback through a Feedback Loop (FB), which might guide the LLM's subsequent interactions.
\end{itemize}

Based on the presented workflow, we now elucidate the decision-making process of the Language Learning Model (LLM) based on two primary scenarios: one where the LLM interacts with the Knowledge Graph (KG) to deduce answers and another where the LLM operates autonomously. 

\subsubsection{Interactions between LLM and KG}

The LLM's reliance on KG is made manifest in the following application scenarios:

\textbf{1. Similar Article Finder:}

For implementing recommendation systems in media and journalism services, it is imperative to identify articles akin to a given one. Using the Cosine similarity metric, we ascertain the resemblance by evaluating the cosine value between the focal article and others. The topmost articles in terms of similarity scores are then selected, as represented in Algorithm \ref{Top 5 Similar Articles}.

\begin{algorithm}
\SetAlgoLined
\KwData{Graph database comprising articles}
\KwResult{Five articles most similar to Article 100}
\BlankLine
\textbf{Procedure:}\\
\Begin{
    \textbf{Query Execution:}\\
    \Indp
    \texttt{MATCH (a1:Article \{article\_id: 100\}), (a2:Article)\\
    WHERE a1 <> a2\\
    WITH a1, a2, gds.similarity.cosine\\
    (a1.content\_vector, a2.content\_vector) AS similarity\_score\\
    RETURN similarity\_score, a1.article\_id, a2.article\_id\\
    ORDER BY similarity\_score DESC LIMIT 5};
    \Indm
}
\caption{Identify Top 5 Articles resembling Article 100}
\label{Top 5 Similar Articles}
\end{algorithm}

\textbf{2. Article Sentiment Analysis:}

In situations necessitating sentiment extraction from a specific article or specific articles with the same set of sentiments, the model retrieves the sentiment, pre-recorded in the article node within the KG. This retrieval is achieved by executing a MATCH Cypher query based on the article's unique identifier, as delineated in Algorithm \ref{Retrieve Sentiment}.

\begin{algorithm}
\SetAlgoLined
\KwData{Graph database containing articles}
\KwResult{Sentiment associated with article\_id 100}
\BlankLine

\textbf{Initialization:}\\
Establish a connection to the graph database.\\
\BlankLine

\textbf{Procedure:}\\
\Begin{
    \textbf{Query Execution:}\\
    \Indp
    \texttt{MATCH (n:Article) WHERE n.article\_id = 100}\\
    \texttt{RETURN n.sentiment}\\
    \Indm
    \BlankLine
    \textbf{Result Display:}\\
    Display the fetched sentiment value.\\
}

\textbf{Termination:}\\
Terminate the connection to the graph database.\\

\caption{Sentiment Extraction for Article with article\_id 100}
\label{Retrieve Sentiment}
\end{algorithm}

\textbf{3. Article Topic Prediction:}

Articles devoid of any designated topic can be assigned one based on the topics of closely resembling articles. This is realized by associating the given article with others and then linking the most similar one's topic. Algorithm \ref{Retrieve One Similar Article} elucidates the related Cypher query.

\begin{algorithm}
\SetAlgoLined
\KwData{Graph database inclusive of articles}
\KwResult{Anticipated topic for article 100 by drawing from its nearest counterpart}
\BlankLine

\textbf{Procedure:}\\
\Begin{
    \textbf{Query Execution:}\\
    \Indp
    \texttt{MATCH (a1:Article \{article\_id: 100\}), (a2:Article)}\\
    \texttt{WHERE a1 <> a2}\\
    \texttt{WITH a1, a2, gds.similarity.cosine(\\
    a1.content\_vector, a2.content\_vector) AS similarity\_score}\\
    \texttt{WHERE similarity\_score > 0.97}\\
    \texttt{RETURN a1.article\_id, a2.article\_id AS similar\_article, t.name AS predicted\_topic, similarity\_score LIMIT 1}\\
    \Indm
}

\caption{Topic Inference for Article 100 from its analogous article}
\label{Retrieve One Similar Article}
\end{algorithm}

\subsubsection{Autonomous LLM Operations}

The LLM, trained on vast datasets, can address certain user queries independently of external databases like the KG. This capability is evident in scenarios requiring comprehension, summarization, or responses that don't necessitate explicit fact-checking against a structured knowledge source.

\textbf{1. Text Summarization:}

For article summarization, the LLM utilizes its training to extract and condense key points without needing KG interaction.

\textbf{2. Generic Journalism Queries:}

LLM can handle queries related to journalistic standards, writing styles, or general media ethics, drawing from its extensive training on journalism-related topics.

\textbf{3. Contextual Interpretations:}

The LLM can interpret and respond to journalism-related queries based on previous context without extracting explicit facts, leveraging its inherent understanding of relational data.

\section{Discussions} \label{sec:discussion}
Conversational AI, while advancing rapidly, faces hurdles in balancing linguistic depth with accurate information representation. As highlighted in our comprehensive review through the Large Language Model Explorer (LLMXplorer), which provides a systemic overview of numerous LLMs, linguistic proficiency often comes at the expense of transparency. Knowledge Graphs, while offering factual precision, might fall short in mimicking human conversational fluidity.

Our exhaustive applied analysis of the practical use cases, challenges, and limitations of LLMs across industries underscores their vast potential. Still, it also highlights the need for architectures that can bridge the aforementioned gaps. In response, our work introduces a functional solution architecture that uniquely integrates KGs and LLMs. This system not only stands out in its linguistic capabilities but also ensures factual consistency. With the incorporation of RBAC, we further the cause of data security, restricting users to role-specific information, and thereby fostering trust for real-world and industry-wide use cases.

The application domain of media and journalism, exemplified using rich data from the real-world product (AI NewsHub) platform, serves as both a case study and a validation point. It is a testament to the architecture's robustness and efficiency. Notably, the adaptability of the proposed architecture means it can seamlessly cater to a myriad of use cases and applications, emphasizing its cross-industry-wide relevance. Some pivotal aspects and design choices to underline the significance of the proposed architecture include:

\begin{itemize}
    \item \textbf{Usage of Specialized LLM (Llama-2-Chat)}: Our choice of the LLM variant, specifically fine-tuned for dialogue, caters to enhanced conversational accuracy and quality.

    \item \textbf{Neo4j for Knowledge Graph Navigation}: The architecture uses Neo4j based on its open-sourceness, industry-wide adoption, proficiency in navigating knowledge structures and efficient data extraction via a query language, Cypher.

    \item \textbf{Cypher Validation Layer}: Located between the LLM and Neo4j, this layer validates Cypher queries from the LLM, mitigating unintended data accesses and system vulnerabilities.

    \item \textbf{Feedback Mechanism}: A feedback loop is integrated for iterative refinement of system responses and continuous LLM learning.

    \item \textbf{Algorithmic Approach}: The architecture employs specific algorithms for tasks such as article similarity detection, sentiment analysis, and topic prediction, ensuring transparency and maintainability.
\end{itemize}

\subsection{Future Implications}

The converging point of linguistic depth and factual accuracy, as epitomized by our system, is a promising indication of where conversational AI is headed. It sets a precedent for building architectures that are not only sophisticated in their response generation but also trustworthy in the information they convey. By addressing key challenges and providing tangible solutions, this work paves the way for future technologies that prioritize both efficiency and trustworthiness.

\section{Conclusion} \label{sec:conclusion}

This research elucidates the advancements and challenges in conversational AI, emphasizing the imperative balance between linguistic richness and factual accuracy. By introducing a novel architecture that synergistically integrates Large Language Models and Knowledge Graphs, we provide a tangible solution to the current transparency and trust issues. Our comprehensive analysis of LLMs and their practical applications across various industries further contributes to the understanding and potential of trustworthy conversational AI. As we look ahead, this work not only addresses present challenges but also charts a promising trajectory for future conversational systems that prioritize efficiency, security, and user trust.

\bibliographystyle{unsrt}
\bibliography{paper}

\appendices

\section{Industry-wide LLM usecases} 
\label{app:usecases}           

\begin{table}[hbt!]
\footnotesize
\renewcommand{\arraystretch}{1.2}
\caption{Applications of LLMs in Finance and Risk Management}
\label{tab:llm-applications-finance}
\begin{tabular}{p{8.1cm}}
\hline
\textbf{Personalized Financial Advising}: Enhancing customer interactions by offering bespoke advice and assistance. \\
\hline
\textbf{Intelligent Support Chatbots}: Addressing customer queries and assisting in real-time. \\
\hline
\textbf{NLP in Financial Analysis}: Parsing unstructured financial data for insights. \\
\hline
\textbf{Market Insights and Sentiment Analysis}: Gleaning consumer behaviour and market trends from diverse sources. \\
\hline
\textbf{Risk Assessment and Portfolio Optimization}: Informing financial decisions through historical data analysis and prediction. \\
\hline
\textbf{Regulatory Adherence}: Assisting institutions in maintaining compliance by analyzing regulatory documentation. \\
\hline
\textbf{Fraud Prevention}: Identifying anomalies in transactions to detect potential fraud. \\
\hline
\textbf{Real-time Market Analytics}: Providing current market sentiments and predictions. \\
\hline
\end{tabular}
\end{table}

\begin{table}[hbt!]
\footnotesize
\renewcommand{\arraystretch}{1.2}
\caption{Applications of LLMs in Healthcare}
\label{tab:llm-applications-healthcare}
\begin{tabular}{p{8.1cm}}
\hline
\textbf{Medical Data Synthesis}: Extracting insights from diverse medical datasets for informed decisions. \\
\hline
\textbf{Clinical Decision Assistance}: Offering patient-centric insights for diagnostic and therapeutic interventions. \\
\hline
\textbf{Medical Imaging Insights}: Detecting anomalies in radiological images. \\
\hline
\textbf{NLP in Medical Documentation}: Assisting with transcription, language translation, and information retrieval. \\
\hline
\textbf{Predictive Health Analytics}: Identifying disease risks and facilitating early interventions. \\
\hline
\textbf{Drug Discovery Support}: Assisting in potential drug target identification and efficacy prediction. \\
\hline
\textbf{Patient Education and Virtual Assistance}: Personalizing medical education and support for patients. \\
\hline
\textbf{Real-time Patient Monitoring}: Providing alerts based on patient data analysis. \\
\hline
\end{tabular}
\end{table}

\begin{table}[hbt!]
\footnotesize
\renewcommand{\arraystretch}{1.2}
\caption{Applications of LLMs in Education and e-Learning}
\label{tab:llm-applications-education}
\begin{tabular}{p{8.1cm}}
\hline
\textbf{Customized Learning Paths}: Adapting content to cater to unique student requirements. \\
\hline
\textbf{Intelligent Tutoring Systems}: Offering interactive guidance and adaptive feedback. \\
\hline
\textbf{Content Generation}: Creating and adapting educational resources. \\
\hline
\textbf{Language Learning Support}: Facilitating exercises, translation, and pronunciation guides. \\
\hline
\textbf{Automated Evaluation}: Grading assignments with detailed feedback. \\
\hline
\textbf{Dynamic Assessments}: Generating assessments tailored to student performance. \\
\hline
\textbf{Resource Recommendations}: Suggesting pertinent educational materials. \\
\hline
\textbf{Gamified Learning Experiences}: Creating engaging, interactive scenarios and simulations. \\
\hline
\end{tabular}
\end{table}

\begin{table}[hbt!]
\footnotesize
\renewcommand{\arraystretch}{1.2}
\caption{Applications of LLMs in Customer Service and Support}
\label{tab:llm-applications-customer-service}
\begin{tabular}{p{8.1cm}}
\hline
\textbf{Personalised Recommendations and Upselling:} Analysing customer data to provide personalized product recommendations, suggest complementary items, and support upselling and cross-selling efforts. \\
\hline
\textbf{Sentiment Analysis and Customer Insights:} Analysing customer feedback and social media conversations to extract sentiment, provide actionable insights, and enhance products, services, and customer experience. \\
\hline
\textbf{Self-Service Support and Knowledge Base:} Assisting in creating and maintaining self-service support systems and knowledge bases, generating FAQs, troubleshooting, and offering step-by-step instructions for customers. \\
\hline
\textbf{Voice and Speech Recognition:} Contributing to voice and speech recognition technologies, accurately transcribing and interpreting customer voice inputs, enabling voice-controlled self-service support and interactions. \\
\hline
\textbf{Call Center Support:} Aiding call centre agents by providing real-time suggestions, relevant information, and access to knowledge bases during customer calls, enhancing efficiency and customer satisfaction. \\
\hline
\textbf{Language Translation and Multilingual Support:} Facilitating language translation in customer service interactions, overcoming language barriers, and supporting multilingual customer service. \\
\hline
\textbf{Email and Ticket Management:} Assisting in managing customer support emails and tickets, categorizing inquiries, prioritizing urgent cases, and suggesting appropriate responses. \\
\hline
\textbf{Customer Feedback Analysis:} Analyzing customer feedback surveys, reviews, and ratings to extract insights, understand preferences, and make data-driven decisions for product or service enhancements. \\
\hline
\textbf{Complaint Resolution and Escalation:} Assisting in complaint resolution by providing guidelines, suggested responses, or escalation procedures to customer service agents, ensuring consistent and appropriate handling. \\
\hline
\end{tabular}
\end{table}

\begin{table}[hbt!]
\footnotesize
\renewcommand{\arraystretch}{1.2}
\caption{Applications of LLMs in Marketing and Sales}
\label{tab:llm-applications-marketing-sales}
\begin{tabular}{p{8.1cm}}
\hline
\textbf{Content Generation:} Assisting in generating marketing content, suggesting ideas, and automating parts of the content creation process. \\
\hline
\textbf{Personalised Marketing Campaigns:} Analysing customer data to create personalised marketing campaigns and tailored messages based on demographics and preferences. \\
\hline
\textbf{Market Research and Trend Analysis:} Assisting in market research by analysing industry reports, customer reviews, and social media conversations to identify trends, preferences, and competitive insights. \\
\hline
\textbf{Customer Profiling and Segmentation:} Analysing customer data to generate detailed customer profiles and segment customers based on various attributes. \\
\hline
\textbf{Social Media Listening and Engagement:} Monitoring social media platforms, tracking brand mentions, engaging with customers, and responding to inquiries or comments. \\
\hline
\textbf{Lead Generation and Qualification:} Analysing customer data to identify potential leads, assess lead quality, predict purchase intent, and prioritise leads for follow-up. \\
\hline
\textbf{Sales Support and Product Information:} Providing sales teams with real-time product information, pricing details, and assistance in answering customer queries and objections. \\
\hline
\textbf{Customer Journey Mapping and Optimization:} Analysing customer interactions, feedback, and behavioural data to map the customer journey, identify pain points, and optimise the customer experience. \\
\hline
\textbf{Marketing Automation and Personalization:} Powering marketing automation platforms, segmenting audiences, crafting personalised campaigns, and delivering dynamic content based on user preferences. \\
\hline
\end{tabular}
\end{table}

\begin{table}[hbt!]
\footnotesize
\renewcommand{\arraystretch}{1.2}
\caption{Applications of LLMs in Human Resources and Recruitment}
\label{tab:llm-applications-hr}
\begin{tabular}{p{8.1cm}}
\hline
\textbf{Resume Screening and Candidate Evaluation:} Analysing and evaluating candidate qualifications, skills, and experience against job requirements, shortlisting candidates based on predefined criteria. \\
\hline
\textbf{Job Description Generation:} Generating comprehensive job descriptions based on industry trends, suggesting appropriate job titles, responsibilities, and qualifications. \\
\hline
\textbf{Candidate Sourcing and Talent Pooling:} Searching through candidate data to identify potential candidates based on specific skills, experience, or qualifications, assisting in building talent pools. \\
\hline
\textbf{Interview Assistance:} Providing interview guidance, suggesting questions, evaluation criteria, and offering insights into candidate background information. \\
\hline
\textbf{Employee Onboarding and Orientation:} Supporting employee onboarding by providing resources, and policies, and generating onboarding materials such as welcome emails and orientation guides. \\
\hline
\textbf{HR Policy and Compliance Guidance:} Assisting with HR policy guidance, answering HR-related questions, interpreting employment laws, and ensuring compliance with regulations and standards. \\
\hline
\textbf{Employee Engagement and Performance Management:} Aiding in employee engagement and performance management by generating surveys, collecting feedback, and providing insights for improvement. \\
\hline
\textbf{HR Knowledge Base and Self-Service Support:} Contributing to HR knowledge bases, generating responses to HR questions, and offering guidance on policies, procedures, benefits, and programs. \\
\hline
\textbf{Employee Performance Analysis:} Analysing employee performance data, identifying patterns, strengths, areas for improvement, and development opportunities, aiding in performance evaluations. \\
\hline
\textbf{Employee Satisfaction Surveys:} Generating employee satisfaction surveys, analysing collected data, and providing insights to improve employee engagement, satisfaction, and retention. \\
\hline
\textbf{Employee Training and Development:} Contributing to employee training initiatives, generating training materials, courses, and resources to enhance effectiveness and scalability. \\
\hline
\textbf{Diversity and Inclusion Initiatives:} Supporting diversity and inclusion efforts by analysing employee demographics, suggesting strategies, and generating diversity reports. \\
\hline
\textbf{Employee Exit Interviews and Feedback Analysis:} Automating exit interviews, analysing employee feedback, identifying areas of improvement, and providing insights for addressing concerns and retention. \\
\hline
\textbf{HR Analytics and Predictive Modelling:} Analysing HR data to derive insights, predict trends, and assist in workforce planning, succession planning, and talent management strategies. \\
\hline
\textbf{Employee Benefits and Wellness Programs:} Providing information and guidance on employee benefits, wellness initiatives, and resources to promote employee well-being. \\
\hline
\textbf{HR Compliance and Ethics:} Assisting with HR compliance, providing information on employment laws, anti-discrimination policies, and ethical guidelines. \\
\hline
\end{tabular}
\end{table}

\begin{table}[hbt!]
\footnotesize
\renewcommand{\arraystretch}{1.2}
\caption{Applications of LLMs in Legal and Compliance}
\label{tab:llm-applications-legal}
\begin{tabular}{p{8.1cm}}
\hline
\textbf{Legal Research and Case Analysis:} Assisting in legal research, extracting insights from legal documents, case law, statutes, and regulatory materials, and supporting attorneys in case analysis and preparation. \\
\hline
\textbf{Contract Analysis and Due Diligence:} Reviewing contracts, extracting key terms, conditions, obligations, and risks, aiding in due diligence processes, and identifying potential legal issues. \\
\hline
\textbf{Compliance Monitoring and Risk Assessment:} Monitoring for regulatory compliance, risk assessment, and suggesting mitigative actions based on analysis of internal processes and external regulations. \\
\hline
\textbf{Legal Document Generation:} Assisting in the generation of legal documents, contracts, agreements, and other legal materials by suggesting standardized language and clauses. \\
\hline
\textbf{E-Discovery and Document Review:} Aiding in e-discovery processes, categorizing, and tagging legal documents, identifying relevant information, and optimizing the document review process. \\
\hline
\textbf{Legal Knowledge Management:} Assisting in organizing and accessing legal knowledge resources, suggesting appropriate references, and facilitating the retrieval of information. \\
\hline
\textbf{Legal Compliance Training and Support:} Generating training materials for legal compliance, offering guidance on regulatory requirements, and assisting in creating compliance programs. \\
\hline
\textbf{Legal Language Translation and Interpretation:} Assisting in translating legal documents and terminology across languages, supporting multilingual legal practices, and ensuring accurate interpretation. \\
\hline
\textbf{Legal Chatbots and Virtual Assistants:} Providing general legal guidance, answering common legal queries, and offering initial consultation through virtual platforms and chatbots. \\
\hline
\textbf{Dispute Resolution and Mediation Support:} Assisting legal professionals in dispute resolution and mediation processes, suggesting strategies, providing relevant case law, and offering insights. \\
\hline
\end{tabular}
\end{table}

\begin{table}[hbt!]
\footnotesize
\renewcommand{\arraystretch}{1.2}
\caption{Applications of LLMs in Supply Chain and Manufacturing}
\label{tab:llm-applications-supply}
\begin{tabular}{p{8.1cm}}
\hline
\textbf{Supply Chain Optimization:} Assisting in supply chain optimization efforts, suggesting strategies, analysing logistics and distribution networks, and providing insights for streamlined operations. \\
\hline
\textbf{Quality Control and Defect Detection:} Assisting in quality control processes by detecting anomalies in production data, identifying potential product defects, and suggesting corrective actions. \\
\hline
\textbf{Predictive Maintenance:} Analysing equipment and machinery data, predicting potential failures, suggesting maintenance schedules, and optimizing equipment uptime. \\
\hline
\textbf{Demand Forecasting and Inventory Management:} Analysing sales data, historical demand, and other external factors to predict product demand, aiding in inventory optimization, and reducing stock-outs or overstock situations. \\
\hline
\textbf{Supplier Relationship Management:} Evaluating supplier performance, identifying risks or inefficiencies, and providing insights for better supplier collaboration and relationship management. \\
\hline
\textbf{Product Design and Development:} Offering insights for product design based on market trends, customer feedback, and competitive analysis, aiding in product innovation and development. \\
\hline
\textbf{Sustainability and Environmental Impact:} Analysing supply chain data to suggest environmental sustainability initiatives, reduce carbon footprint, and align operations with sustainability goals. \\
\hline
\textbf{Supply Chain Risk Management:} Identifying potential risks in the supply chain, suggesting risk-mitigative strategies, and ensuring consistent supply chain operations. \\
\hline
\textbf{Production Planning and Scheduling:} Assisting in production planning, suggesting optimized schedules based on demand forecasts, resource availability, and operational constraints. \\
\hline
\textbf{Lean Manufacturing and Process Improvement:} Analysing manufacturing processes, identifying inefficiencies, suggesting lean strategies, and providing insights for process improvement and optimization. \\
\hline
\end{tabular}
\end{table}

\end{document}